\documentclass[10pt,journal,compsoc]{IEEEtran}

\ifCLASSOPTIONcompsoc
  \usepackage[nocompress]{cite}
\else
  \usepackage{cite}
\fi

\ifCLASSINFOpdf
\else
\fi
\usepackage{amsmath}
\usepackage{mathrsfs}
\usepackage{amsfonts,amssymb}
\usepackage{amsthm}
\theoremstyle{plain}
\newtheorem{Def}{Definition}

\newtheorem{thm}{Theorem}

\usepackage{subfigure}
\usepackage{array}
\usepackage{graphicx}
\usepackage{booktabs}
\usepackage{epstopdf}

\usepackage{algorithm}
\usepackage{algorithmic}
\usepackage{setspace}

\usepackage{url}

\begin{document}
%
\title{MMCGAN: Generative Adversarial Network with Explicit Manifold Prior}
%
%
%
%

\author{Guanhua~Zheng,
        Jitao~Sang,
        and~Changsheng~Xu
\IEEEcompsocitemizethanks{\IEEEcompsocthanksitem G. Zheng is with University of Science and Technology of China (e-mail: zhenggh@mail.ustc.edu.cn).\protect\\
\IEEEcompsocthanksitem J. Sang and J. Yu are with the School of Computer and Information Technology and the Beijing Key Laboratory of Traffic Data Analysis and Mining, Beijing Jiaotong University, Beijing 100044, China (e-mail: \{jtsang, jianyu\}@bjtu.edu.cn).\protect\\

\IEEEcompsocthanksitem C. Xu is with the National Lab of Pattern Recognition, Institute of Automation, CAS, Beijing 100190, China, and University of Chinese Academy of Sciences(e-mail: csxu@nlpr.ia.ac.cn).
}
}

\IEEEtitleabstractindextext{%
\begin{abstract}
Generative Adversarial Network(GAN) provides a good generative framework to produce realistic samples, but suffers from two recognized issues as mode collapse and unstable training. In this work, we propose to employ explicit manifold learning as prior to alleviate mode collapse and stabilize training of GAN. Since the basic assumption of conventional manifold learning fails in case of sparse and uneven data distribution, we introduce a new target, Minimum Manifold Coding (MMC), for manifold learning to encourage simple and unfolded manifold. In essence, MMC is the general case of the shortest Hamiltonian Path problem and pursues manifold  with minimum Riemann volume. Using the standardized code from MMC as prior, GAN is guaranteed to recover a simple and unfolded manifold covering all the training data. Our experiments on both the toy data and real datasets show the effectiveness of MMCGAN in alleviating mode collapse, stabilizing training, and improving the quality of generated samples.
\end{abstract}

}

\maketitle

\IEEEdisplaynontitleabstractindextext

%
\IEEEpeerreviewmaketitle
\section{Introduction}
\label{introduction}
Generative Adversarial Networks (GANs) \cite{goodfellow2014generative} can generate outstanding realistic examples, but suffer from two recognized problems (illustrated in Fig.1(a),(b) respectively):
(1) Mode collapse, the generators of GANs could commonly miss modes in the training data while successfully cheating the discriminators; (2) Training instability, the training processes of GANs may fail at certain stages of the training. Many researchers have been devoted to solving them theoretically \cite{arjovsky2017wasserstein,zhao2016energy,salimans2018improving} or empirically ~\cite{salimans2016improved,metz2016unrolled,lin2018pacgan}, but these problems are still open.

In this work, we tackle these problems by imposing explicit manifold prior onto GANs.
GAN is recognized to model manifold from observed samples~\cite{zhu2016generative}. Since no latent representation are explicitly provided for the observed samples, GAN can be seen as implementing implicit manifold learning. It is worth noting that explicit manifold learning has advantages correspondingly addressing the above two problems: (1) By explicitly coding each observed sample on the generated manifold, all modes in the training data are guaranteed to be recovered and thus mode collapse problem can be solved naturally; (2)  Explicit manifold learning has effect of pulling the generated manifold to the observed samples, which provides effective gradient to avoid the training instability at the beginning of training~\cite{salimans2016improved}.

However, directly employing the conventional manifold learning methods fails to recover the intrinsic manifold to generate realistic samples. Manifold learning methods usually follow  important assumptions like neighbors points lying close to a locally linear patch or preserving local structures, which are generally not satisfied in case of sparsely or unevenly distributed data. To construct an appropriate prior for GAN, we are motivated to further simplify the generated manifold to address the shortage in training data. Specifically, a new target for manifold learning, \emph{Minimum Manifold Coding}(MMC), is imposed to encouraging small Riemann volume of the generated manifold. The proposed MMC turns out a general form of the Shortest Hamiltonian Path(SHP) problem~\cite{polychronopoulos1996stochastic}, which aims to find a minimum manifold with fixed dimensions to cover all the samples and thus guarantees the simple and crease-free generated manifold. The standardized codes derived from MMC are then employed as prior to regularize the generator training in GAN, which constitutes the proposed framework of MMCGAN. The generated samples from MMCGAN in addressing mode collapse and training instability problems are illustrated in Fig.1(c),(d) correspondingly. We have conducted experiments on both toy datasets of 2D-SwissRoll, 25-Grid and realistic datasets of MNIST, Cifar10, ImageNet to show the
 effectiveness of MMC and MMCGAN. The main contributions can be summarized in three-fold:
\begin{itemize}
\item We propose to employ explicit manifold learning as prior to address the mode collapse and training instability problems of GAN.
\item A new manifold learning target of Minimum Manifold Coding (MMC) is imposed to tackle the sparse and uneven data distribution and provide more suitable prior for GAN training. An approximate solution is also provided for the MMC problem.
\item Extensive experiments shows that MMCGAN can alleviate mode collapse, stabilize training, and improve the quality of generated samples on different GAN architectures.
\end{itemize}

\begin{figure*}[t]
\centering
\subfigure[WGAN-gp] {
\includegraphics[width=1.5in]{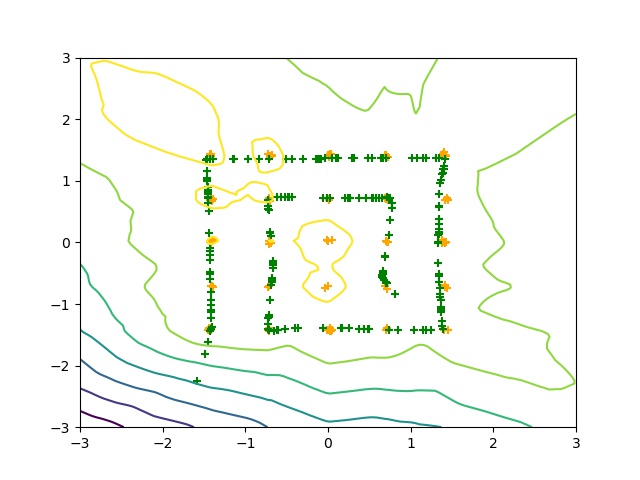}
}
\subfigure[standard GAN] {
\includegraphics[width=1.5in]{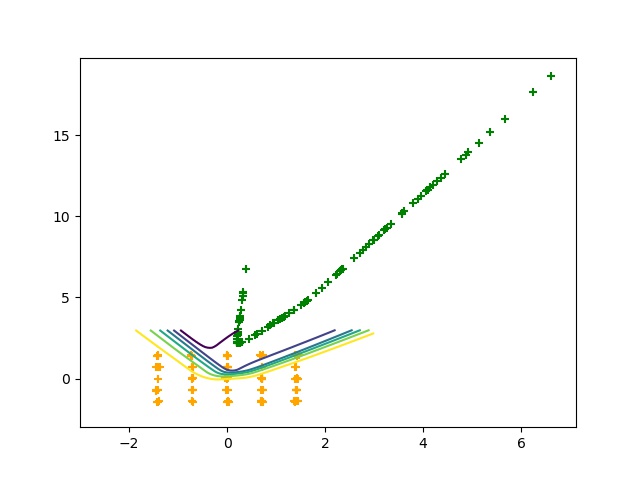}
}
\subfigure[MMC+WGAN-gp] {
\includegraphics[width=1.5in]{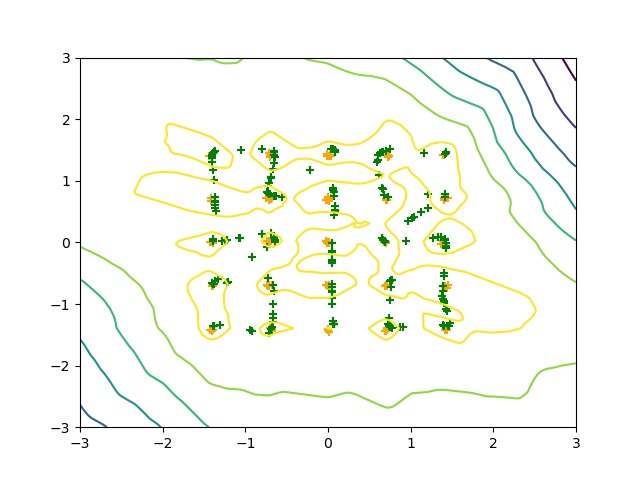}
}
\subfigure[MMC+standard GAN] {
\includegraphics[width=1.5in]{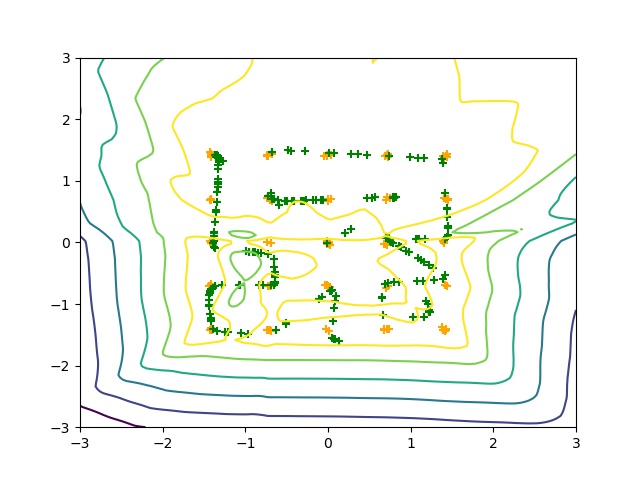}
}
\caption{Examples of (a) mode collapse and (b) training instability on the 25-Grid dataset. (c) and (d) shows the corresponding results from the proposed MMCGAN.}
\end{figure*}

\section{Related Work}
\label{sec2}
\subsection{Manifold Learning}
\label{sec2.3}
Manifold learning assumes that data are distributed around some low-dimensional manifolds rather than all over the data spaces.  The goal of manifold learning is to discover this low-dimensional compact representation for the high-dimensional data. Classical manifold learning methods include LLE~\cite{roweis2000nonlinear}, Isomap~\cite{tenenbaum2000global}, Laplacian Eigenmaps~\cite{belkin2003laplacian} ,
ltsa~\cite{zhang2004principal}, t-SNE~\cite{maaten2008visualizing}, LargeVis~\cite{tang2016visualizing}, Umap~\cite{mcinnes2018umap}, etc. These methods basically consist of three steps: (1) finding k-nearest neighbors; (2) constructing a graph to preserve the structures of the raw data; and (3) embedding the raw data into low-dimensional representation satisfying the manifold structure.

While manifold learning methods are widely used in data visualization and dimensionality reduction problems, they are not readily used as prior for generating realistic samples. One of the most important reasons is that manifold learning assume a topological space where every point has a neighborhood that is homeomorphic to the interior of a sphere in
Euclidean space. Therefore, k-nearest neighbors represent the local structures only if the data are dense and evenly-distributed, which is hardly satisfied for realistic samples like
 image datasets. This critically limits the integration of conventional manifold learning methods into generative models like GAN. In this work, we introduce a further MMC target to simplify the generated manifold and fit to GAN in generating realistic samples. Comparison results with conventional manifold learning methods will be reported in the experiment section.

\subsection{Generative Adversarial Networks}
\label{sec2.1}
Generative Adversarial Networks have two major parts: Generator (G) and Discriminator (D). The original form of GAN ~\cite{goodfellow2014generative} aims to find a Nash equilibrium
to the following min-max problem:
\begin{equation}\label{rawgan}
\min_{G}{\max_{D}{\mathbb{E}_{x \sim q_{data}}[\log{D(x)}]+\mathbb{E}_{z \sim p_{z}}[\log{(1-D(G(z)))}]}}
\end{equation}
where $z \sim R^m$ is a latent representation drawn from distribution $p_z$ such as $\mathcal{N}(0,1)$ or $\mathcal{U}[1,-1]$.

Theoretically, at the global optimum of Eqn.(\ref{rawgan}), generator will produce samples with the same distribution as data distribution. Unfortunately, standard GAN does not work
well as it tends to be unstable during training, and its generator may treat the discriminator without diversities, which is called mode collapse.

Many researchers make efforts to solve these problem \cite{nowozin2016f,fedus2017many,salimans2018improving,mescheder2018training}. An important line of work start from
WGAN~\cite{arjovsky2017wasserstein}, which provides comprehensive theoretical analysis and gained good experimental performance. WGAN theoretically analyzes the reason why GAN is unstable and solve it by using
Wasserstein distance to substitute Jensen-Shannon divergence in GAN. Then, the objective function changes to:
\begin{equation}\label{wgan}
\min_{G}{\max_{D}{\mathbb{E}_{x \sim q_{data}}[D(x)]+\mathbb{E}_{z \sim p_{z}}[-D(G(z))]}}
\end{equation}
Note that the discriminators here needs to be $1$-$Lipstchitz$. WGAN achieves this target by clipping the gradients. After that, WGAN-gp\cite{gulrajani2017improved} provides a more stable
 solution by imposing gradient penalty: the derived gradients will not be limited in only two values $\{-1,1\}$. SNGAN\cite{miyato2018spectral} is the state-of-art choice in this line of work,
which is faster than WGAN-gp and achieves better performance. Recently, SNGAN was been implemented in BigGAN \cite{brock2018large}, which is the first photo-realistic GAN with the following hinge loss as objective function:
\begin{equation}\label{sngan}
\begin{split}
&\min_{D}{\mathbb{E}_{x \sim q_{data}}[(1-D(x))_+]+\mathbb{E}_{z \sim p_{z}}[(1+D(G(z)))_+]}\\
&\min_{G}{\mathbb{E}_{z \sim p_{z}}[-D(G(z))]}
\end{split}
\end{equation}
where $(\cdot)_+=\max(\cdot,0)$. We will compare the proposed MMCGAN with these typical GAN architectures to examine the effectiveness in addressing mode collapse and training stability.

\subsection{GAN with Reconstruction Loss}
\label{sec2.2}
In this work, manifold learning serves as prior for GAN by adding a manifold preserving reconstruction loss during training the generator. In fact, reconstruction loss is one intuitive and efficient way to guarantee GAN not to lose information. This subsection reviews some GAN variants with reconstruction loss to penalizing losing different types of information. CycleGAN\cite{zhu2017unpaired} employs reconstruction loss to constrain that the image-to-image translation generates more diverse
samples. EBGAN \cite{zhao2016energy} replaces traditional discriminator loss with reconstruction loss to show the performance of other energy functions. BAGAN \cite{yang2017dagan} uses the reconstruction loss to training a decoder as a better initialization for the generator, however, such initialization is a trick without theoretically analysis, and the coding of auto-encoder have a completely different distribution from $p_z$, which make its benefit unobvious. The most similar study is sinGAN ~\cite{shaham2019singan}, which uses reconstruction loss to guarantee that there exists an input noise to generate each of the raw image samples. The difference lies in that, sinGAN is proposed for a single input, which will not converge in case of multiple samples with random noise. Moreover, the reconstruction loss introduced in this work is motivated from manifold preserving perspective, which is compatible with the manifold discovery nature of generative models.

\section{Minimum Manifold Coding}
\label{sec4}
As discussed in Introduction, in case of sparse and uneven data distribution, it is difficult for the generator to correctly recover manifold and generate realistic samples. We are motivated to introduce a new manifold learning target, Minimum Manifold Coding(MMC), to address these problems. Such target encourages a simple and unfolded manifold, so that the generators can fit easily. In this section,
we will first derive the formal definition of the MMC, and then analyze its correlation with
the Shorted Hamiltonian Path to explain why MMC leads to unfolded manifold. Finally, we will provide a practical algorithm to solve MMC.

\subsection{Notations and Definitions}
\label{sec4.1}
\textbf{Notations}. Let the input data be $X=\{x_1,x_2,...,x_N\}$ where $\forall i,x_i \in R^n$, and we suppose all the samples are different from each other.
 The manifold learning methods embed $X$ to a low-dimensional space: we can obtain
 a set of codes corresponding to the input data: $C=\{c_1,c_2,...,c_N\}$, where $\forall i, c_i \in R^m$, $m<n$. Note that such codes represent the encoding mapping: $c_i=C(x_i), \forall i \in \{1,2,...,N\}$.
The decoding function $f_C$ can recover the data $X$ from the coding $C$, and we denote $F_C$ as the set of all the decoding functions of $C$: $F_C=\{f_C:R^m\mapsto R^n| \forall c_i \in C, x_i=f_C(c_i)\}$. In addition, a decoding function $f_C$ can
 generate a corresponding manifold $M(f_C)=\{x|\forall c\in R^m, x=f_C(c)\}$. Obviously, all these manifolds of decoding functions in $F_C$ will intersect at the raw data points $X$.

Recall that our target is to find a simple and unfolded manifold so that the generator of GAN can fit easily. Intuitively, the manifold with the minimum Riemann volume is simple, and we derive a new objective named Minimum Manifold Coding(MMC) from this motivation. Before deriving the formal definition of MMC, we will define \emph{Mapping Measure} first.

 \begin{Def}\label{MM}
(Mapping Measure).
Let the convex hull for a coding $C$ is $S=conv(C)$. A decoding function $f_C$ maps $S$ to the corresponding manifold: $f_C(S)\subset M(f_C)$.
The mapping measure for $f_C$ defined as the Riemann volume of $f_C(S)$:
\begin{equation}
\Lambda(f_C)=\int_{S}{\sqrt{\left | det(J_{f_C}(s)^TJ_{f_C}(s)) \right|}ds}
\end{equation}
Where $det$ is the determinant of the matrix, and $J_{f_C}$ is the Jacobian matrix of $f_C$.
\end{Def}
 \begin{Def}\label{MMC}
(Coding Measure, Coding Manifold, and Minimum Manifold Coding).
Let $F_C$ be a set of all the decoding functions with the coding $C$. The coding measure of the coding $C$ is the minimum mapping measure of the functions in $F_C$:
\begin{equation}
\rho(C)=\min_{f_C\in F_C} {\Lambda(f_C)}
\end{equation}

The coding manifold is the manifold generated by the decoding function $f_C$ with the minimum mapping measure:
\begin{equation}
M(C)=M(\arg\min_{f_C\in F_C} {\Lambda(f_C)})
\end{equation}

Minimum manifold coding is to find a coding $C'$ with the minimum coding measure:
\begin{equation}
C'= \arg\min_{C}{\rho(C)}
\end{equation}
\end{Def}

\begin{figure}[t]
\vskip 0.2in
\begin{center}
\centerline{\includegraphics[width=0.7\columnwidth, height=1.2in]{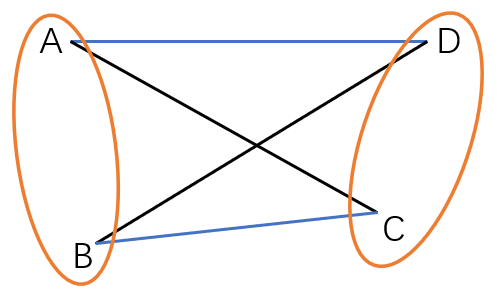}}
\caption{Illustration of the shortest Hamiltonian Path with no cross.}
\end{center}
\vskip -0.2in
\end{figure}
\subsection{General Form of the Shortest Hamiltonian Path}
Such definition of MMC has a good property: the coding measure only depends on the arrangement of the data on the manifold, rather than the scale of codes. For example, if $m=1$,
the manifolds will be curves, i.e., 1-D manifolds. As coding measure is the minimum mapping measure, it represents a set of line segments which connects
all the points. Specifically, we can visualize the coding manifolds with painting line segments from the data point with the minimum code to the point with the maximum code in the data space. It is clear that such manifold only depends on the order of the codes, rather than the specific code values.

As a matter of fact, the minimum manifold coding with 1-D manifold is equivalent to the Shortest Hamiltonian Path(SHP). Since the convex hull for a 1-D coding $C$ is a line segment ranging from the minimum code to the maximum code, the coding measure of the coding $C$ represents a path visiting each vertex exactly once, which is called Hamiltonian Path~\cite{tutte1946hamiltonian}.
Therefore, minimizing the manifold coding can retrieve the shortest Hamiltonian path. In other words, MMC can be seen as a general form of the shortest Hamiltonian path.

It is worth noting that the shortest Hamiltonian path represents a simple curve with less cross. Suppose there is a cross ($AC+BD$) in Hamiltonian path, see Fig.2. Without loss of generality, we suppose the cross aims to connect $AB$ and $CD$. It is clear that $AD+BC$ can also achieve the same connection target with no less, and $AD+BC<AC+BD$. As the general form of SHP, MMC is expected to discover manifold with less cross or even unfolded.

\subsection{Approximate Solution of MMC}
\label{sec4.3}
As known to all, the SHP is an NP problem, so the MMC problem is also an NP problem and can only be approximately solved. In this subsection, we provide a practical approximate solution of the MMC problem.
In brief, we split this problem into two parts: getting the decoding functions, and pursuing smaller mapping measures. For the first part, we use an auto-encoder with reconstruction loss. For the second part, we have the following theorem:
\begin{thm}\label{thm}
Let $f_C\in F_C$ be a decoding function which satisfies the L-Lipschitz condition on $S=conv(C)$, then the mapping measure of $f_C$ has an upper bound:
\begin{equation}
\Lambda(f_C)=\int_{S}{\sqrt{\left | det(J_{f_C}(s)^TJ_{f_C}(s)) \right|}ds}\leq L^m \int_{S}{ds}
\end{equation}
where $m$ is the dimension of the coding space.
\end{thm}
The proof is provided in Supplement-A. According to this theorem, as there always exist an $L$ to make decoder satisfy the $L$-$Lipschitz$ condition,
we can use a minimum convex hull loss to get smaller convex hull $S$ and obtain a lower upper bound. In this work, we choose the constraint of L2-Norm for simplicity, so the objective function for auto-encoder becomes:
\begin{equation}\label{gam}
\min_{Dec,Enc}{\frac{1}{2}\mathbb{E}_{x \sim q_{data}}(\|x-Dec(Enc(x))\|^2+\gamma\|Enc(x)\|^2)}
\end{equation}
where $Dec$ is the decoder, and $Enc$ is the encoder of the auto-encoder. After training, we can obtain a coding with small coding measure. Recall that the coding measure will not change if we use the transformation which do not alter the arrangements of coding, so we can design a proper transformation to obtain an expected coding distribution. Note that the latent representations of GAN is drawn from the distribution $\mathcal{N}(0,1)$, and a code with zero-mean and one-variance will be more reasonable as the prior.  In this work, we use z score standardization as the transformation function: $C'=\frac{C-\mathbb{E}C}{std(C)}$.
\begin{figure}[t]
\vskip 0.2in
\begin{center}
\centerline{\includegraphics[width=0.65\columnwidth, height=1.4in]{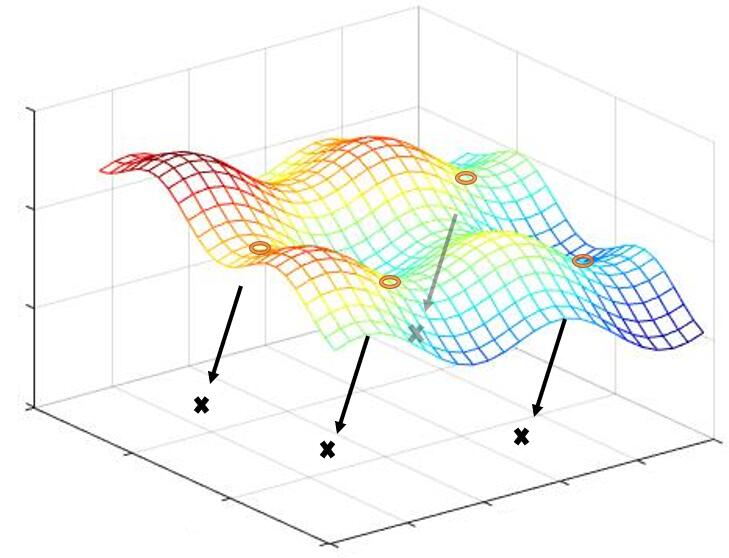}}
\caption{The manifold preserving reconstruction loss pulls the generator manifold to the data points according the corresponding codes.}
\label{mc}
\end{center}
\vskip -0.2in
\end{figure}
\begin{figure*}[t]
\vskip 0.2in
\begin{center}
\centerline{\includegraphics[width=5.5in]{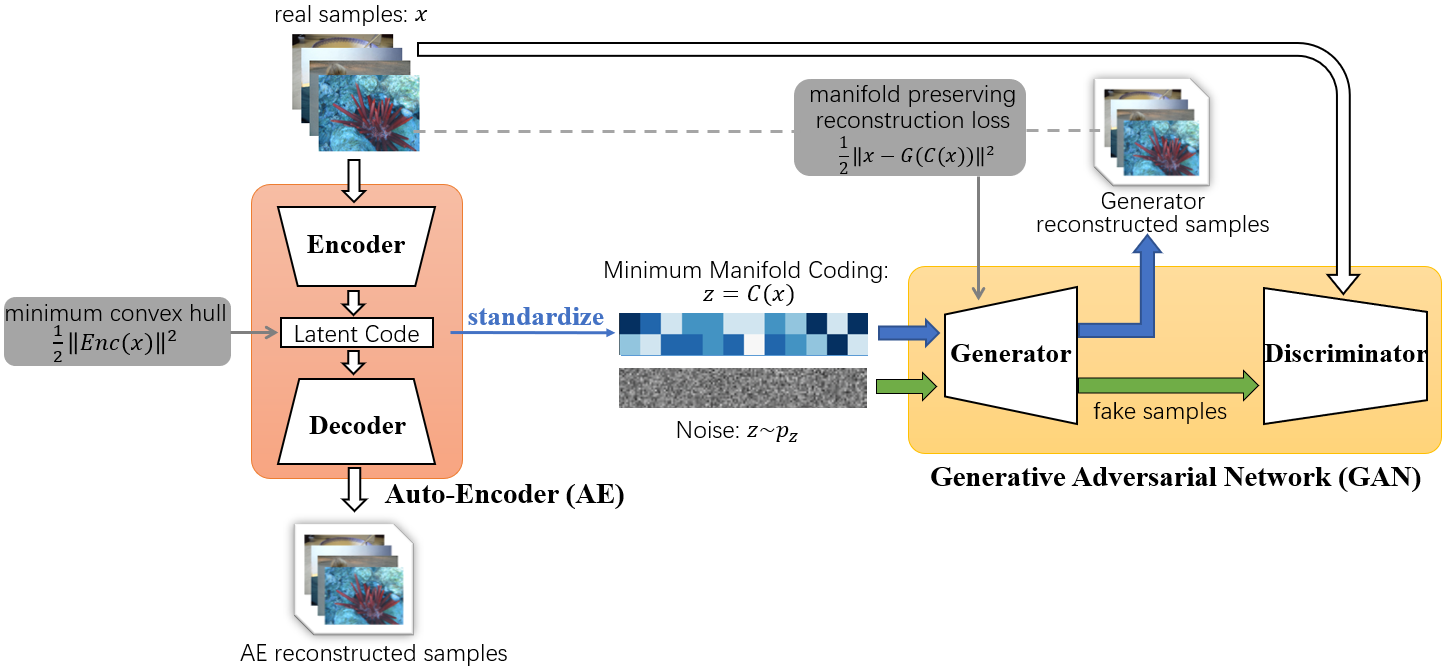}}
\caption{Illustration of MMCGAN framework.}
\end{center}
\vskip -0.2in
\end{figure*}
\section{MMCGAN}
\label{sec3.1}
In this section, we will use the manifold obtained by MMC as prior to improve the training of GAN, so called MMCGAN.
There are many ways to implement the idea of using explicit manifold learning prior. We employ one intuitive way to constrain the generator to fit the prior manifold by $L2$ loss between the generator manifold and prior manifold: $R=\mathbb{E}_{x \sim q_{data}}\|x-G(C(x))\|^2$. We call the L2 loss as manifold preserving reconstruction loss.

The overall framework of MMCGAN is shown in Fig.4, which consists of the auto-encoder component to derive the manifold prior, and the GAN component to using the prior to regularize the generator training. The training process has three steps: in the first step, we employ auto-encoder with loss of convex hull to get the latent code that minimizes the mapping measure. The derived code is then standardized by with z-score to be compatible with the input distribution of GAN.

In the second step, we use the standardized code as prior to initialize the generator. Specifically, manifold preserving reconstruction loss is imposed at the beginning of the generator training, e.g., for hinge loss, we have:
\begin{equation}\label{rlhinge}
\begin{split}
&\min_{D}{\mathbb{E}_{x \sim q_{data}}[(1-D(x))_+]+\mathbb{E}_{z \sim p_{z}}[(1+D(G(z)))_+]}\\
&\min_{G}{\mathbb{E}_{z \sim p_{z}}[-D(G(z))]}+\frac{\lambda}{2}\mathbb{E}_{x \sim q_{data}}\|x-G(C(x))\|^2
\end{split}
\end{equation}
Fig.3 illustrates the role of the manifold preserving reconstruction loss: it can be seen as some anchor pulling the generator manifold close to the data manifold, and ensure GANs are capable of producing all the input samples. Furthermore, GAN training is usually unstable at the beginning because of the adversarial mechanism. The manifold preserving reconstruction loss can provide consistent gradients to stabilize training.

When the generator manifold is close enough to the AE recovered manifold, the role of manifold preserving reconstruction loss will be trivial. In contrast, further imposing the loss will prevent the generator from exploring its potential to cheat the discriminator. Moreover, the Nash equilibrium is difficult to achieve and guarantee the generator distribution is the same as the data distribution. Therefore, in the third step, we remove the manifold preserving reconstruction loss and turn to the training GAN in the standard way. Empirically, we use the moving average of the reconstruction loss to measure the closeness between the generator manifold and AE recovered manifold. A threshold value $T$ is set, and the third step switches on when the moving average is below $T$.
\begin{figure*}[t]
\centering
\subfigure[MMC] {
\includegraphics[width=1.2in]{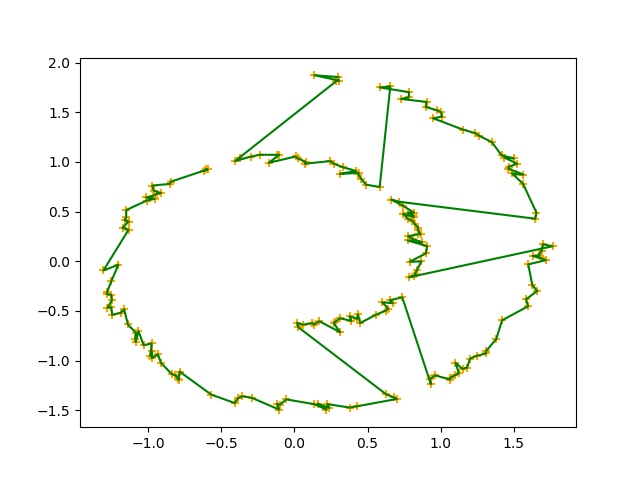}
}
\subfigure[Laplace Eigenmap] {
\includegraphics[width=1.2in]{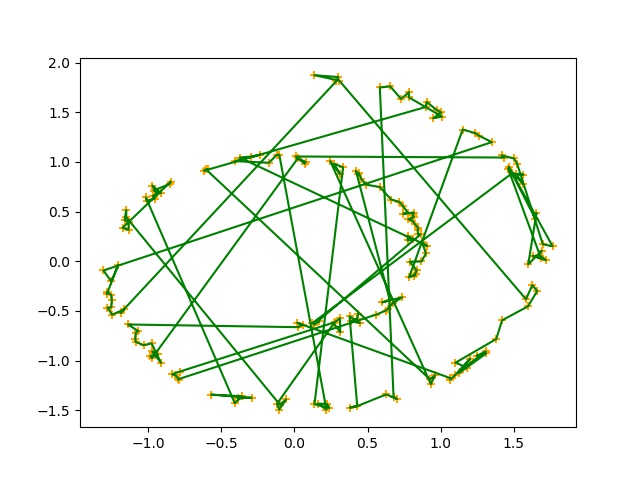}
}
\subfigure[LLE] {
\includegraphics[width=1.2in]{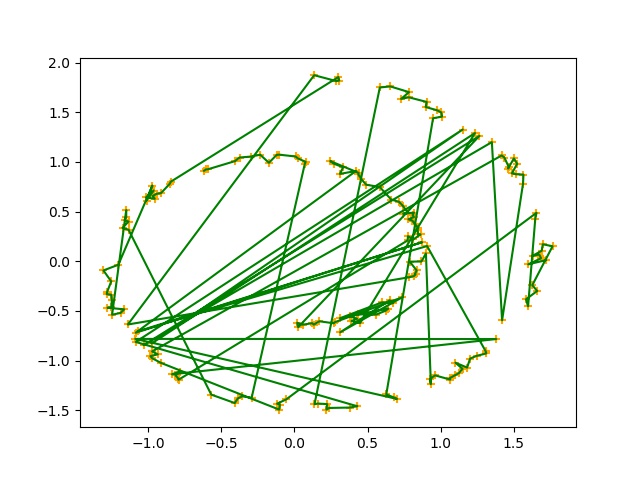}
}
\subfigure[ltsa] {
\includegraphics[width=1.2in]{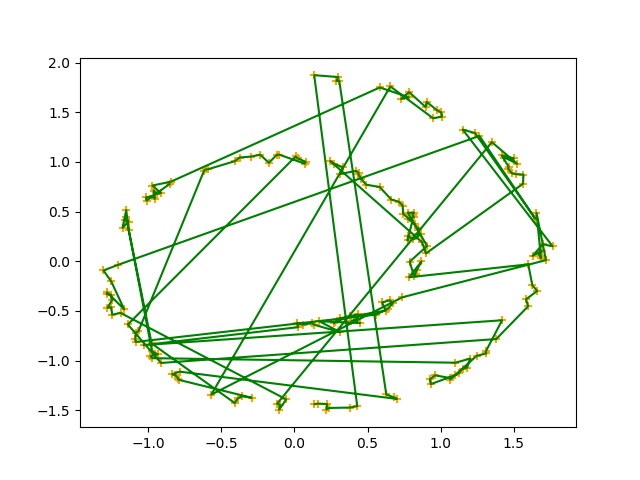}
}
\subfigure[UMAP] {
\includegraphics[width=1.2in]{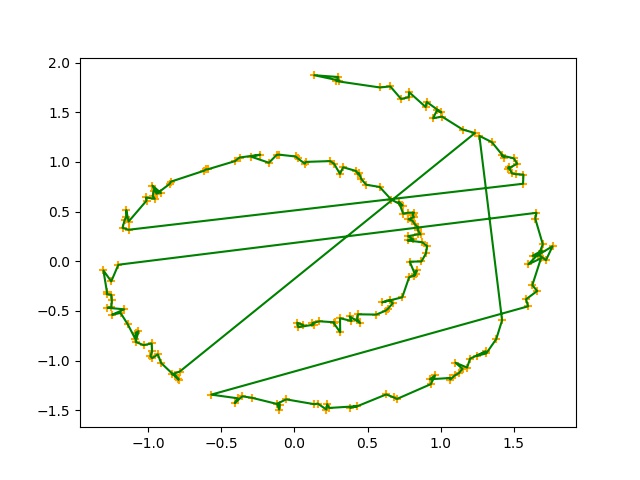}
}
\subfigure[MMC] {
\includegraphics[width=1.2in]{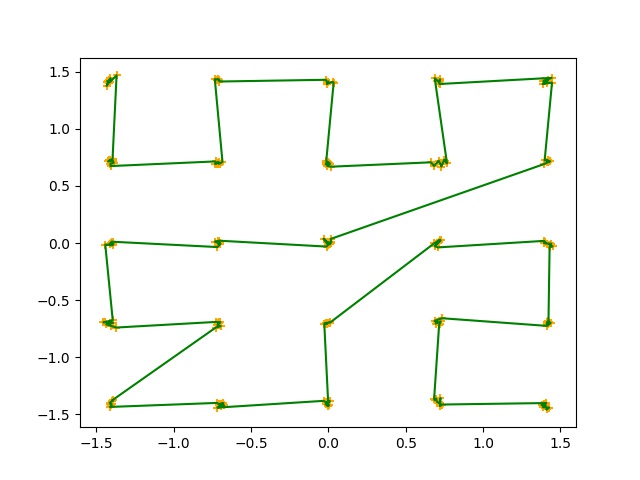}
}
\subfigure[Laplace Eigenmap] {
\includegraphics[width=1.2in]{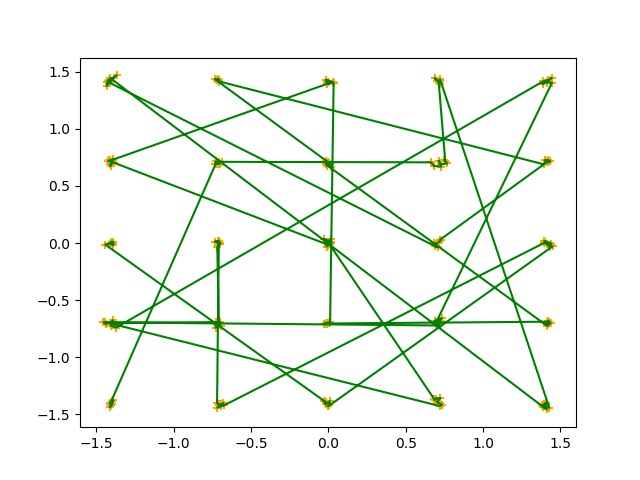}
}
\subfigure[LLE] {
\includegraphics[width=1.2in]{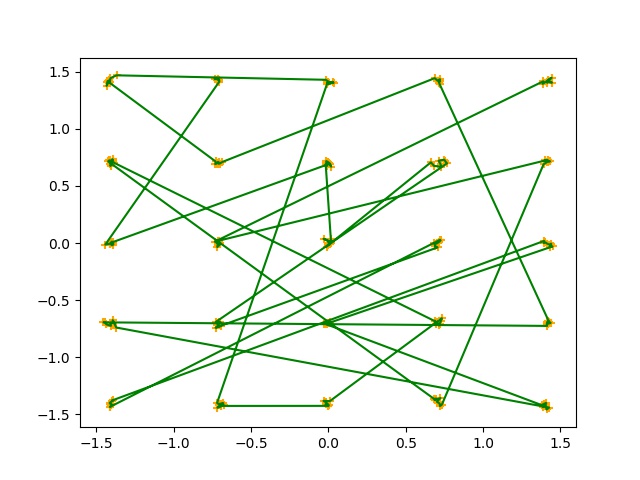}
}
\subfigure[ltsa] {
\includegraphics[width=1.2in]{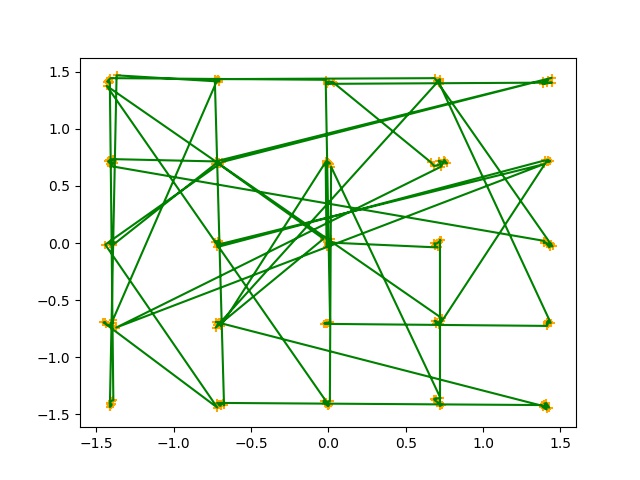}
}
\subfigure[UMAP] {
\includegraphics[width=1.2in]{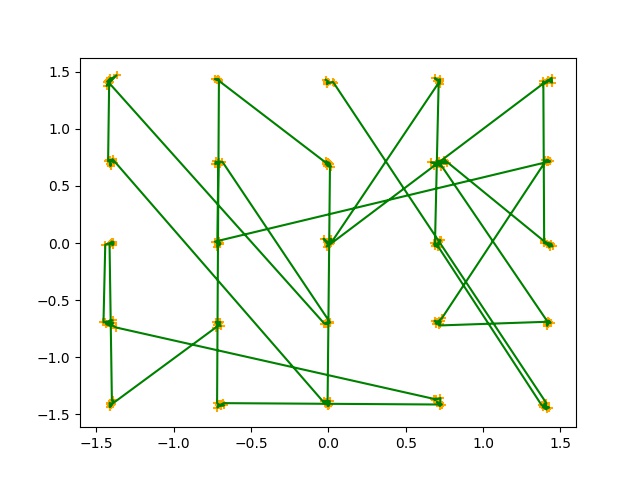}
}

\caption{Recovered manifolds by  MMC and traditional manifold learning methods on: (a)-(e) 2D-SwissRoll and (f)-(i) 25-Grid.}
\label{fig2}
\end{figure*}

\section{Experiments}
\label{sec5}
In this section, after introducing the experiment settings of MMCGAN, we first compare MMC with
 the traditional manifold learning on sparse and uneven data, and then evaluate MMCGAN on different datasets with widely used GAN architectures to show its effectiveness in avoiding mode collapse and stabilizing training.

\subsection{Implementation Settings}
Note that delicate tuning of model hyperparameters and learning parameters is not necessary for MMCGAN, as most settings are universal for different datasets and model architectures. For the model hyperparameters, there are two hyper-parameters: we report the experimental results using $\gamma=\frac{1}{10m}$ for Eqn.(\ref{gam}) and $\lambda=1$ for Eqn.(\ref{rlhinge}), where $m$ is the dimension of latent representations in GAN.

For the learning parameters, we enumerate them according to the training process illustrated in Section 4.
For the first step, we use Adam optimizer~\cite{kingma2014adam} with $\beta_1 = 0.5$, and  $\beta_2 = 0.9$. The learning rate scheme is described in SGDR\cite{loshchilov2016sgdr}
 which can accelerate convergence, with $T_0=10$, $\eta_{min}=0$ and $\eta_{max}=0.001$.
For the second step, the moment of moving average is $0.999$ in this work, and the threshold $T$ is the moving average of the reconstruction loss of the first step. We  have conducted experiments on 5 datasets and the following list the choice of threshold for these experiments: (1) \emph{2D-SwissRoll}, 0.1; (2)\emph{25-Grid}, 0.01; (3) \emph{MNIST}, 6; (4) \emph{Cifar10}: 30; (5) \emph{ImageNet20}: 1000.
For the third step, the training settings are all the same as the normal GAN. The specific hyperparameters and architectures of benchmark GANs used in practice are detailed in Supplement-B.

In addition, all the experiments use data-parallel distributed training in Pytorch with 6 Nvidia Titan X 12G. The source codes can be obtained in the supplementary materials.

\subsection{MMC Evaluation}
\label{sec5.1}
 The choice of explicit manifold learning prior determines the performance. To evaluate the performance of MMC prior, we conducted experiments on two synthetic datasets which are sparsely and unevenly distributed respectively: (1) \emph{2D-SwissRoll}, 200 samples which obtained by \verb|sklearn.datasets.make_swiss_roll| with noise of 0.25. To make the results more clear, we only use the first two dimensions and scale down to $\frac{2}{15}$. (2) \emph{25-Grid}\cite{lin2018pacgan}, 200 data samples from a mixture of 25 two-dimensional Gaussians with the same variances $\frac{1}{3200}$ and different means $(\frac{i}{2\sqrt{2}},\frac{j}{2\sqrt{2}})$, where $i, j \in \{-2, -1, 0, 1, 2\}$.

We also examined the traditional manifold learning methods: PCA~\cite{wold1987principal}, Isomap~\cite{tenenbaum2000global}, Laplacian Eigenmaps~\cite{belkin2003laplacian}, LLE~\cite{roweis2000nonlinear}, HLLE~\cite{donoho2003hessian}, MLLE~\cite{zhang2007mlle}, ltsa~\cite{zhang2004principal}, t-SNE~\cite{maaten2008visualizing}, and Umap~\cite{mcinnes2018umap}.
These methods are all implemented with official packages or sklearn, and the hyperparameters are \verb|n_neighbors=3, n_components=1|.

To evaluate the performance intuitively, as analyzed in Section 3.2, we can use a set of line segments to connect all the data according to the order of coding to show the performance of the manifold learning. Fig.5 shows the results of MMC and 4 of the traditional manifold learning methods. Results for other manifold learning methods can be seen in Supplement-C. It can be seen that the manifold recovered by the examined traditional manifold learning methods tend to be folded and twisted, while MMC derives simple manifolds based on the proposed approximate solution.

\subsection{MMCGAN Evaluation}

\begin{figure}[t]
\vskip 0.2in
\begin{center}
\centerline{\includegraphics[width=0.7\columnwidth]{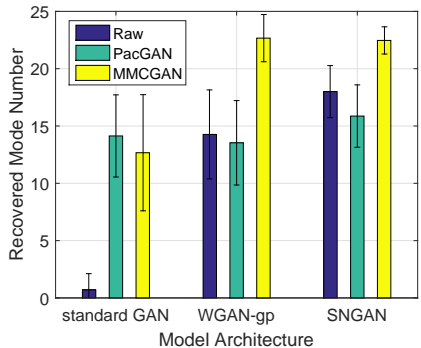}}
\caption{Mode collapse experiments: comparing between standard GAN architectures, PacGAN and the proposed MMCGAN.}
\label{mc}
\end{center}
\vskip -0.2in
\end{figure}
\begin{figure*}[t]
\centering
\subfigure[WGAN-gp] {
\includegraphics[width=1.6in]{gaussian_wgan.jpg}
}
\subfigure[WGAN-gp(PacGAN)] {
\includegraphics[width=1.6in]{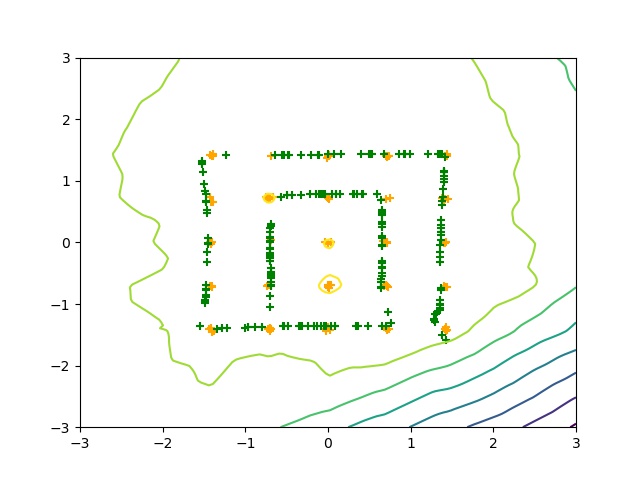}
}
\subfigure[MMC+WGAN-gp] {
\includegraphics[width=1.6in]{gaussian_wganour.jpg}
}

\subfigure[SNGAN] {
\includegraphics[width=1.6in]{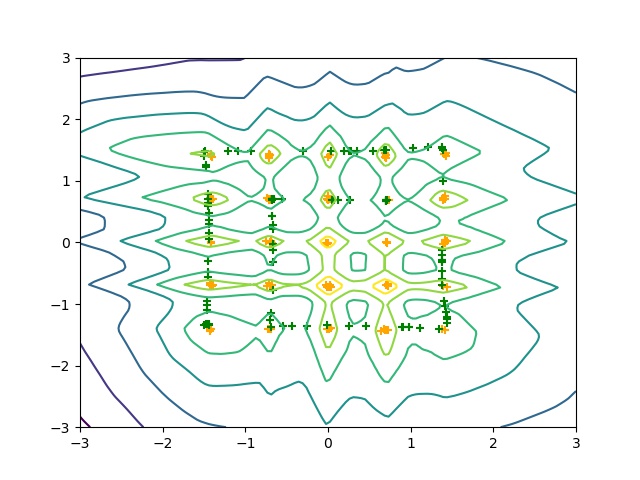}
}
\subfigure[SNGAN(PacGAN)] {
\includegraphics[width=1.6in]{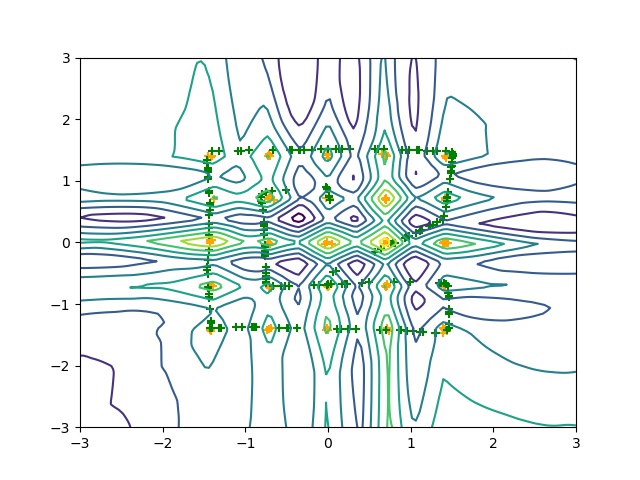}
}
\subfigure[MMC+SNGAN] {
\includegraphics[width=1.6in]{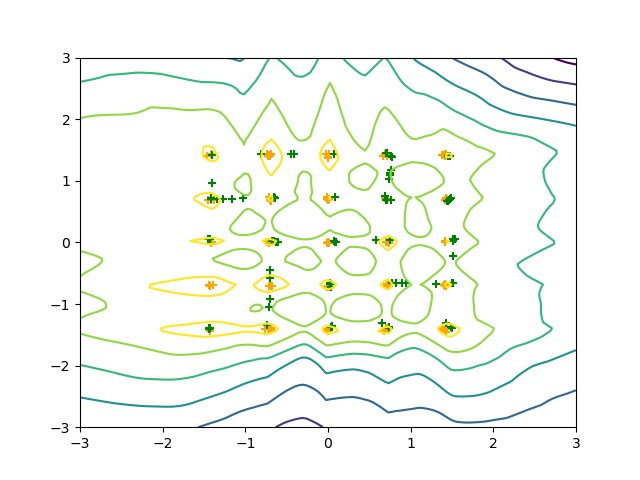}
}
\caption{Intuitive results of mode collapse for the WGAN-gp and SNGAN architectures. }
\label{fig}
\end{figure*}

\subsubsection{Mode Collapse Results}
\label{sec5.2}
Mode collapse indicates that the generator only produces data within a subset of modes. Recently, such phenomenon is not well-understood, and previous work provide many hypothesis,
like improper objective functions~\cite{arora2017generalization,arjovsky2017wasserstein} and weak discriminators~\cite{li2017towards,salimans2016improved}. Based on these hypothesis, previous work have proposed many methods, e.g., ATI~\cite{dumoulin2016adversarially}, VEEGAN~\cite{srivastava2017veegan}, unrolled GAN~\cite{metz2016unrolled}. The state-of-art is PacGAN\cite{lin2018pacgan}, which strengthens the discriminators by packing the inputs. In this subsection, we compare MMCGAN with PacGAN on \emph{25-Grid} on three different architecture: standard GAN, WGAN-gp, SNGAN.

Specifically, we set the latent representation with dimension $m=1$ to make the generator produce a low-dimensional manifold. As \emph{25-Grid} is constructed by 25 Gaussian distributions, it has 25 different modes, and we use the distance between the samples and the centers of 25 Gaussian distributions to examine whether the generator can produce such modes. Specifically, we sampled from the generator distributions 200 times, and recorded how many modes have samples near enough, i.e., the distance is less than $0.1$.
All the experiments repeat 5 times. In each experiment, to obtain a stable results, we averaged the last 5 results before the end of training.

 The results are summarized in Fig.6. It can be seen that MMCGAN can obtain more modes in \emph{25-Grid} datasets for WGAN-gp and SNGAN, but work slightly poorer for the standard
 GAN. That is because MMCGAN can only improve the initialize states of training, while the standard GAN has recognized problem of its global minima~\cite{salimans2016improved}, which is well solved in WGAN-gp and SNGAN. PacGAN also contains special mechanism to address this problem by strengthening the discriminator. The intuitive results of WGAN-gp and SNGAN architectures can be seen in Fig.7, MMCGAN successfully cover almost all the modes while the raw GAN and PacGAN usually miss some modes especially in the sparsely and unevenly distributed area.

\begin{figure}[t]
\centering
\subfigure[standard GAN] {
\includegraphics[width=1.5in]{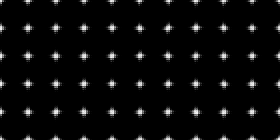}
}
\subfigure[MMC+standard GAN] {
\includegraphics[width=1.5in]{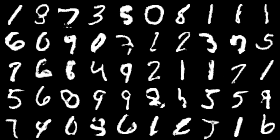}
}
\subfigure[SNGAN] {
\includegraphics[width=1.5in]{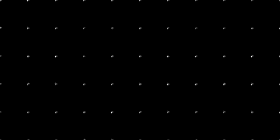}
}
\subfigure[MMC+SNGAN] {
\includegraphics[width=1.5in]{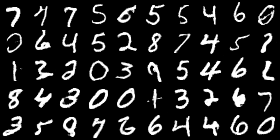}
}
\caption{Visualization of samples produced by generators on MNIST: (a) standard GAN without BatchNorm; (c) SNGAN without BatchNorm; (b) and (d) illustrate the corresponding results with MMC prior.}
\label{fig}
\end{figure}
\subsubsection{Training Stability Results}
\label{sec5.1}
We use two datasets to show the performance of MMCGAN in stabilizing training:

(1) \emph{25-Grid}:
We visualize the generator distributions in Fig.1(b)(d),  where the green points are the fake data and the yellow points are the training data. We also paint the contour lines of the discriminators to show the training trend. It can be seen that GAN with standard objective (Eqn.(\ref{rawgan})) is very fragile: the generator manifold deviates too far to
 fit the data. The proposed MMC prior successfully avoid such deviation for stable training.

(2) \emph{MNIST}~\cite{lecun1998mnist}:
we use the training set which consists of 60K $28\times28$ images of handwritten digits. The benchmark architecture is 3-layer DCGAN \cite{radford2015unsupervised} without BatchNorm~\footnote{Batch normalization plays an important role in stabilize training of DCGAN, we remove it to obtain an unstable control group to show the effect of MMCGAN.}~\cite{ioffe2015batch}. The generated images are visualized in Fig.8: standard GAN(Eqn.(\ref{rawgan})) and SNGAN with hinge loss (Eqn.(\ref{sngan})) without BatchNorm both failed, and adding MMC prior successfully recovered the data manifold and generated the realistic handwritten digit images.
\begin{figure}[t]
\centering
\subfigure[IS] {
\includegraphics[width=1.5in]{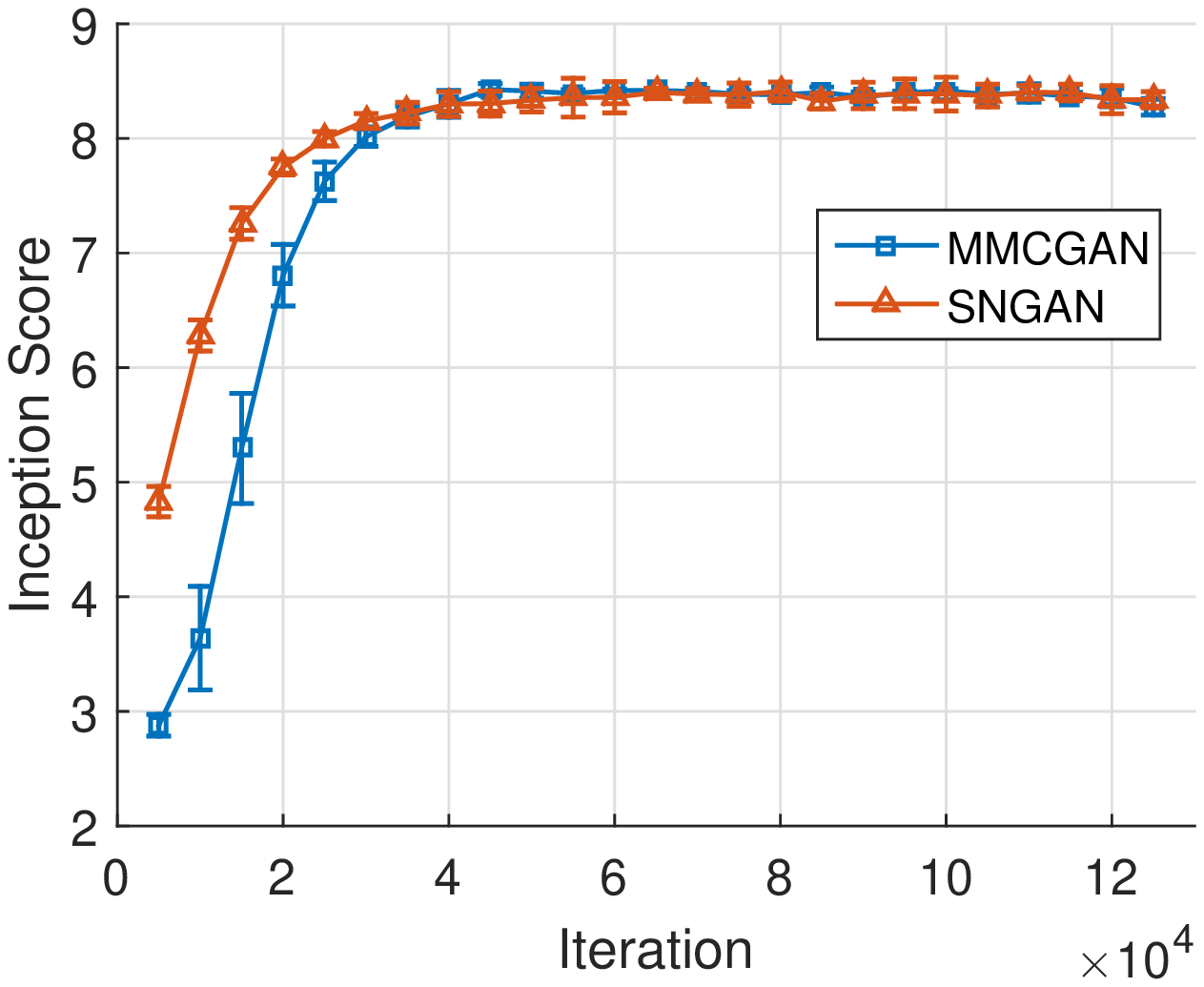}
}
\subfigure[FID] {
\includegraphics[width=1.5in]{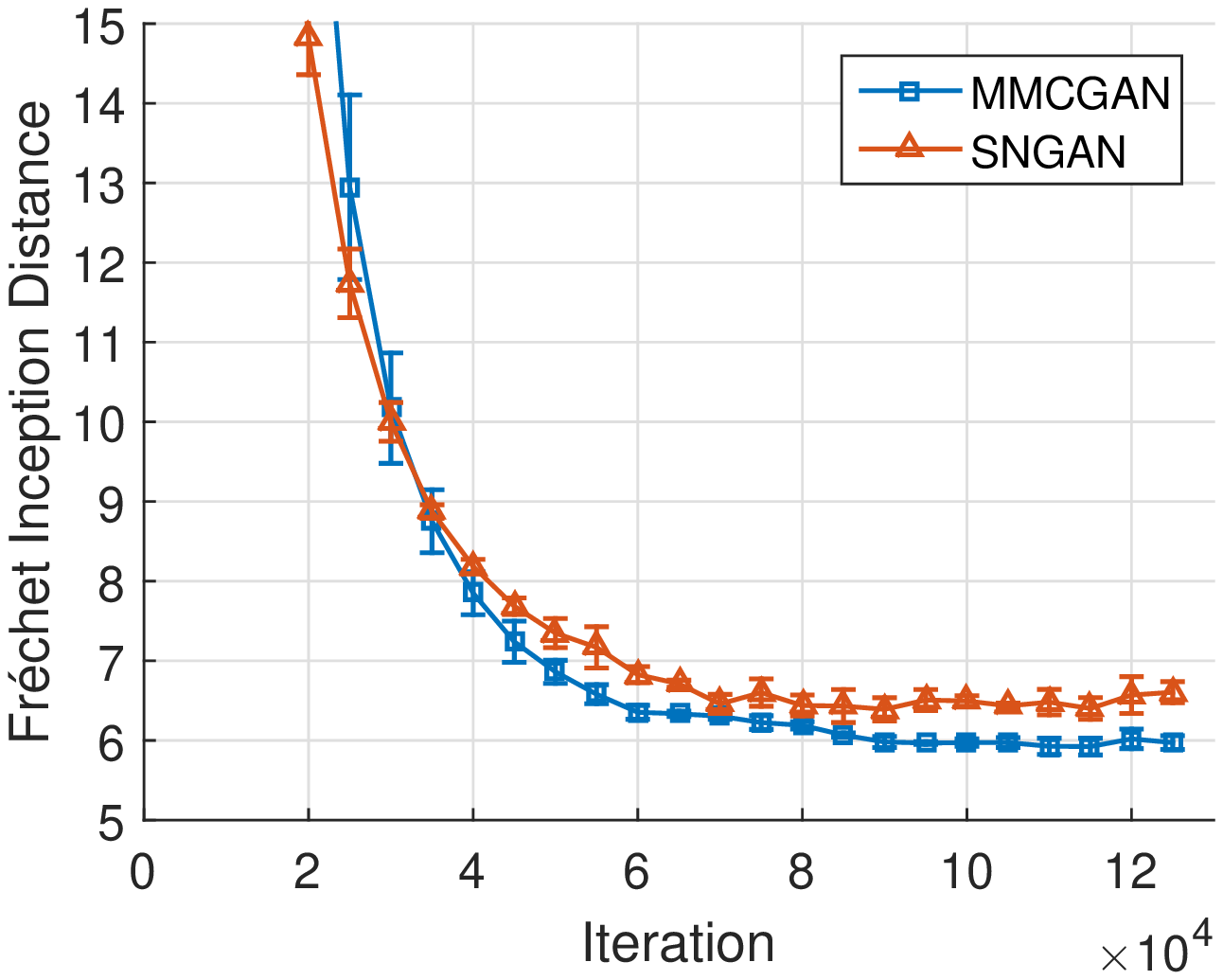}
}
\caption{Inception score and FID of MMCGAN and baseline on Cifar10. }
\end{figure}

\begin{figure}[t]
\centering
\subfigure[IS] {
\includegraphics[width=1.5in]{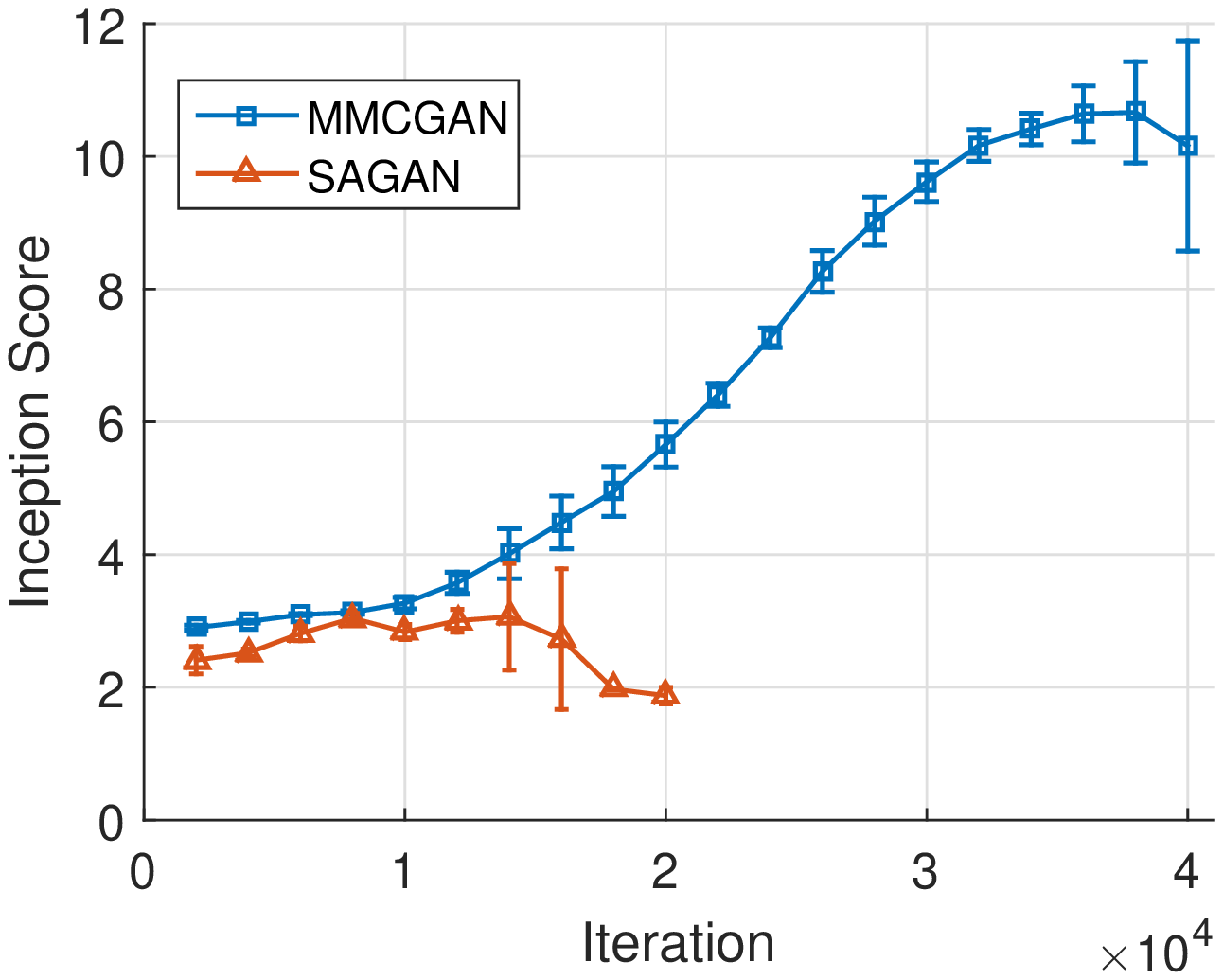}
}
\subfigure[FID] {
\includegraphics[width=1.5in]{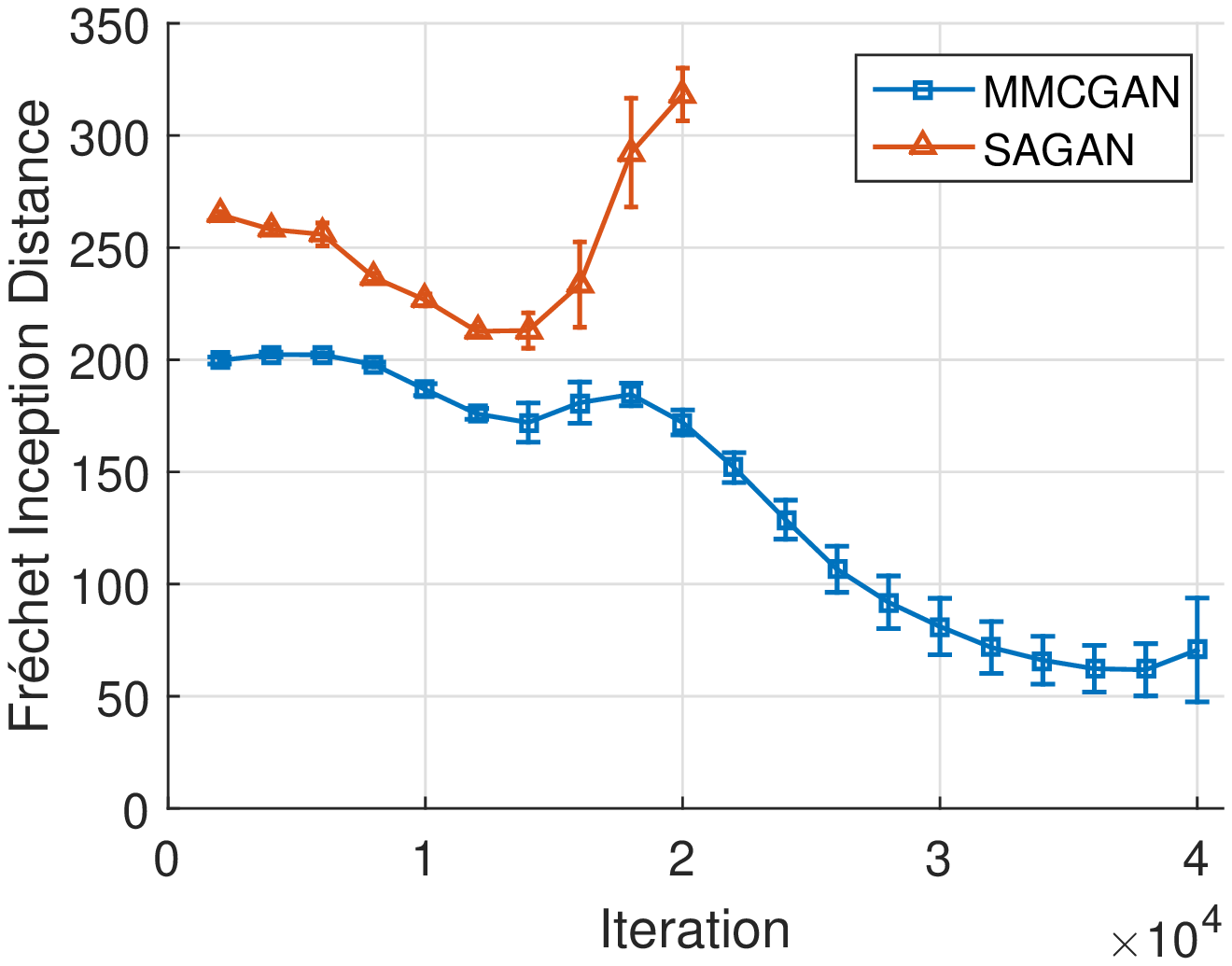}
}
\caption{Inception score and FID of MMCGAN and baseline on ImageNet20.  }
\end{figure}
\subsubsection{Quantitative Results}
\label{sec5.3}
In this subsection, we will examine Inception Score~\cite{salimans2016improved} and FID \cite{heusel2017gans} to quantitatively evaluate the quality of generated samples of MMCGAN. Experiments are conducted on the CIFAR-10 and ImageNet20.

\emph{Cifar-10}: The CIFAR-10 dataset consists of 60k 32$\times$32 color images in 10 classes. We use 50k training images for the training of GAN.
In particular, we choose SNGAN with the implementation of BigGAN as the benchmark.
We report IS and FID measures in Fig.9: As training proceeds,  MMCGAN can improve FID measure while keeping the similar IS measure. This validates that MMCGAN avoid mode collapse and keeps the same global minima.

\emph{ImageNet20}: We select a subset of ImageNet ILSVRC 2012~\cite{deng2009imagenet} for evaluation: 20 categories which start with 'n014' and 'n015', totally about 26k $128\times128$ images. SAGAN~\cite{zhang2018self}
selected as the baseline due to its efficiency in large-scale high resolution datasets. The model was implemented based on the code of BigGAN.

 With each experiment repeating 3 times and calculating the means and standard deviations, Fig.10 plots the error bars for IS and FID measures. All the experiments of traditional SAGAN have broken down before $2\times 10^{4}$ iterations, and the MMCGAN can achieve better performance and keep stable until $4\times 10^{4}$ iterations. Note that the training of SAGAN is stable in the complete ImageNet. The observed instability might be due to the lack of data with such high resolution, where MMC prior successfully stabilize the training process to make up for the data shortage.
\section{Conclusion and Future Work}
\label{sec6}
In this work, we introduce explicit manifold learning as prior for GAN to avoid mode collapse and stabilize training. A new target of Minimum Manifold Coding is further imposed for manifold learning. Such target is validated to discover simple and unfolded manifolds even when the data is sparsely or unevenly distributed.
There remain many interesting directions to be explored in the future. The first direction is the theoretical proof of equilibrium, convergence and analysis on the improvement of mode collapse and training stability. The second direction can  pursue more characteristics of generative models from the perspective of manifold learning, e.g., regularizing the completeness of manifold to obtain balanced  distribution of GAN for data augmentation. Another interesting direction is to explore the potential of GAN beyond data generation. As the generator manifold can closely approach the data manifold with minimum Riemann volume, we can employ GAN to approximate the  solution of MMC, SHP and other similar optimization problems.

\bibliographystyle{IEEEtran}
\bibliography{IEEEabrv,mybib}

\end{document}